\documentclass[10pt,twocolumn,letterpaper]{article}

\usepackage[pagenumbers]{cvpr} 

\usepackage{graphicx}
\usepackage{amsmath}
\usepackage{amssymb}
\usepackage{booktabs}
\usepackage{color}
\usepackage{multirow}
\usepackage{placeins}

\usepackage[table]{xcolor}
\usepackage{pifont}
\usepackage{overpic}
\usepackage{array}
\newcolumntype{P}[1]{>{\centering\arraybackslash}p{#1}}

\definecolor{cvprblue}{rgb}{0.21,0.49,0.74}
\usepackage[pagebackref,breaklinks,colorlinks,citecolor=cvprblue]{hyperref}

\usepackage[capitalize]{cleveref}
\crefname{section}{Sec.}{Secs.}
\Crefname{section}{Section}{Sections}
\Crefname{table}{Table}{Tables}
\crefname{table}{Tab.}{Tabs.}

\def\paperID{3195} 
\def\confName{CVPR}
\def\confYear{2024}

\title{Is Vanilla MLP in Neural Radiance Field Enough for Few-shot View Synthesis?}

\author{Hanxin Zhu\textsuperscript{\rm 1}, Tianyu He\textsuperscript{\rm 2}, Xin Li\textsuperscript{\rm 1}, Bingchen Li\textsuperscript{\rm 1}, Zhibo Chen\textsuperscript{\rm 1}\\
\textsuperscript{\rm 1}University of Science and Technology of China \\
\textsuperscript{\rm 2}Microsoft Research Asia \\
\tt\small hanxinzhu@mail.ustc.edu.cn, tianyuhe@microsoft.com, \\
\tt\small\{lixin666, lbc31415926\}@mail.ustc.edu.cn,  chenzhibo@ustc.edu.cn
}

\begin{document}
\maketitle
\begin{abstract}
Neural Radiance Field (NeRF) has achieved superior performance for novel view synthesis by modeling the scene with a Multi-Layer Perception (MLP) and a volume rendering procedure, however, when fewer known views are given (i.e., few-shot view synthesis), the model is prone to overfit the given views. To handle this issue, previous efforts have been made towards leveraging learned priors or introducing additional regularizations. In contrast, in this paper, we for the first time provide an orthogonal method from the perspective of network structure. Given the observation that trivially reducing the number of model parameters alleviates the overfitting issue, but at the cost of missing details, we propose the multi-input MLP (mi-MLP) that incorporates the inputs (i.e., location and viewing direction) of the vanilla MLP into each layer to prevent the overfitting issue without harming detailed synthesis. To further reduce the artifacts, we propose to model colors and volume density separately and present two regularization terms. Extensive experiments on multiple datasets demonstrate that: 1) although the proposed mi-MLP is easy to implement, it is surprisingly effective as it boosts the PSNR of the baseline from $14.73$ to $24.23$. 2) the overall framework achieves state-of-the-art results on a wide range of benchmarks. We will release the code upon publication.
\end{abstract}    
\section{Introduction}
\label{sec:intro}
Neural Radiance Field (NeRF) has emerged as one of the most promising methods for novel view synthesis, owing to its remarkable ability to represent 3D scenes. By utilizing a Multi-Layer Perception (MLP) in conjunction with classical volume rendering, NeRF can produce photorealistic novel views from multiple 2D images captured from different views~\cite{mildenhall2021nerf}.
Various works extends NeRF to different tasks such as surface reconstruction~\cite{wang2021neus,zhu2023vdn,wang2023pet,verbin2022ref}, dynamic scenes~\cite{pumarola2021d,park2021nerfies,peng2021animatable,gao2021dynamic} and 3D generation~\cite{poole2022dreamfusion,lin2023magic3d,tang2023make,xu2023neurallift,chen2023single,zhou2023sparsefusion}, etc. However, these NeRF-based methods require a large number of input views (\eg, $100$)~\cite{mildenhall2021nerf}. In cases where only a few input views are available (\ie, few-shot view synthesis), NeRF brings severe artifacts and thus leads to a dramatic performance drop~\cite{jain2021putting,roessle2022dense}.

\begin{figure}[t]
    \centering
    \includegraphics[width=1\linewidth]{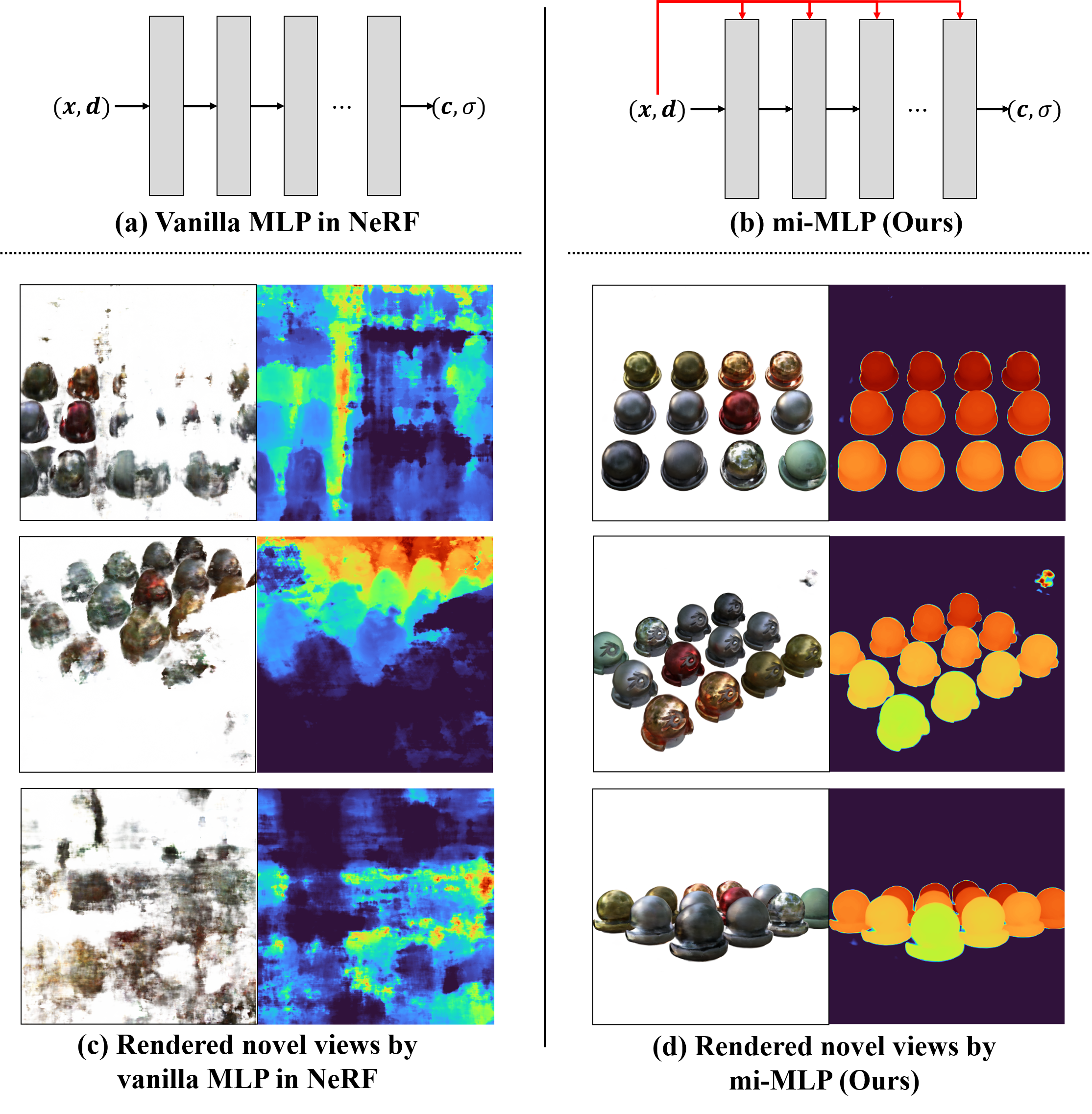}
    \caption{Illustration of \textbf{vanilla MLP vs. mi-MLP}. Although mi-MLP is easy to implement, it is surprisingly effective as it boosts the PSNR of the baseline from $14.73$ to $24.23$.}
    \label{fig: freeze paras}
\end{figure}

Two primary challenges arise in the context of few-shot view synthesis. Firstly, due to the limited amount of training data available, the model is prone to overfitting input views, resulting in the estimated geometry being distributed on 2D planes instead of 3D volumes~\cite{jain2021putting,kim2022infonerf,niemeyer2022regnerf}. Secondly, the presence of artifacts such as ghosting and floating effects significantly limit the fidelity and 3D consistency of rendered novel views~\cite{niemeyer2022regnerf,yang2023freenerf}.

To address the aforementioned issues, mainstream approaches can be categorized into two strategies: prior-based~\cite{chen2021mvsnerf,wang2021ibrnet,yu2021pixelnerf,deng2022depth} and regularization-based~\cite{kim2022infonerf,niemeyer2022regnerf,jain2021putting,yang2023freenerf} methods. Prior-based methods aim to generalize NeRF to different scenes using techniques such as multi-view stereo~\cite{furukawa2015multi} or image-based rendering~\cite{shum2000review}, where a large-scale dataset is utilized to learn scene priors. Regularization-based methods incorporate additional 3D inductive bias, \eg, frequency~\cite{yang2023freenerf} and depth~\cite{niemeyer2022regnerf} regularizations, for the purpose of stronger constraints.
Despite achieving remarkable results, none of these methods take the network structure into account and still adhere to the vanilla MLP~\cite{mildenhall2021nerf}. In this paper, we challenge this common practice and ask: \textit{is vanilla MLP in NeRF enough for few-shot view synthesis}?

To answer this question, we investigate the overfitting issue and have two key observations: 
1) FreeNeRF~\cite{yang2023freenerf} illustrates that the vanilla NeRF is prone to over-fastly converge to high-frequency details. In this way, the model quickly memorizes input views instead of inferring the underlying geometry. Therefore, to avoid overfitting, a direct solution is to decrease the model capacity by reducing the model parameters (\eg, reducing the number of layers); 2) however, as presented in DietNeRF~\cite{jain2021putting}, though the overfitting issue can be alleviated by reducing the model parameters, the details are missed in the generated results. This indicates that model capacity should be preserved for the network.

Capitalizing on the above observations, we propose the multi-input MLP (mi-MLP) that incorporates the inputs (\ie, location and viewing direction) of the vanilla MLP into each layer (as illustrated in Fig.~\ref{fig: freeze paras}). mi-MLP reveals three key insights: \textbf{1)} incorporating the inputs into each layer enables shorter paths between inputs and outputs, allowing synthesis with fewer parameters in an end-to-end way; \textbf{2)} we keep the model capacity unchanged as it is beneficial to synthesizing high-frequency details; \textbf{3)} we keep the inputs and outputs unchanged to make it a plug-and-play solution to the current NeRF-based pipelines.

To further reduce the artifacts, motivated by the assumption that geometry is typically smoother than appearance~\cite{niemeyer2022regnerf}, instead of using a shared model to model the colors and volume density like NeRF, we propose to model them separately to enable positional encoding~\cite{mildenhall2021nerf} with different frequencies. We also propose a novel regularization term to reduce the background artifacts in object-centric scenes and a sampling-annealing strategy to address near-field artifacts in forwarding-facing scenes.

Our main contributions can be summarized as follows:
\begin{itemize}
    \item To address the overfitting issue, we introduce mi-MLP to tackle few-shot view synthesis from the perspective of network structure by incorporating the inputs into each layer.
    \item To achieve better geometry, we propose to model the colors and volume density separately to enable positional encoding with different frequencies.
    \item We propose two regularization terms to improve the quality of rendered novel views.
    \item Through comprehensive experiments, we demonstrate that our method attains superior performance compared with multiple state-of-the-art methods.
\end{itemize}

To the best of our knowledge, this is the first work that tackles NeRF-based few-shot novel view synthesis from the perspective of network structure, opening up a new direction for further research in other fields such as 3D generation.

\section{Related Works}
\label{sec:related works}
\subsection{Neural Radiance Field} 
Neural Radiance Field (NeRF)~\cite{mildenhall2021nerf} has become increasingly popular due to its impressive 3D representation capabilities, where photorealistic novel views can be rendered with 2D posed images.
One of the keys to NeRF's success lies in the usage of an MLP to reason about scene properties, where a mapping from input embeddings to outputs is learned, allowing for continuous scene representation and view interpolation. Numerous researchers have extended NeRF to a variety of areas, including faster training and rendering~\cite{fridovich2022plenoxels,muller2022instant,kerbl20233d}, dynamic scenes~\cite{pumarola2021d,park2021nerfies,peng2021animatable,gao2021dynamic}, generable scenes~\cite{chibane2021stereo,liu2022neural,trevithick2021grf,wang2022generalizable}, and 3D generation~\cite{zeronvs,poole2022dreamfusion,lin2023magic3d}, etc. However, the practical utility of these NeRF-based methods is limited due to the need for a large number of input views. In this paper, we propose a novel method that targets few-shot view synthesis through a well-designed network structure.

\begin{figure*}[t]
    \centering
    \includegraphics[width=1\linewidth]{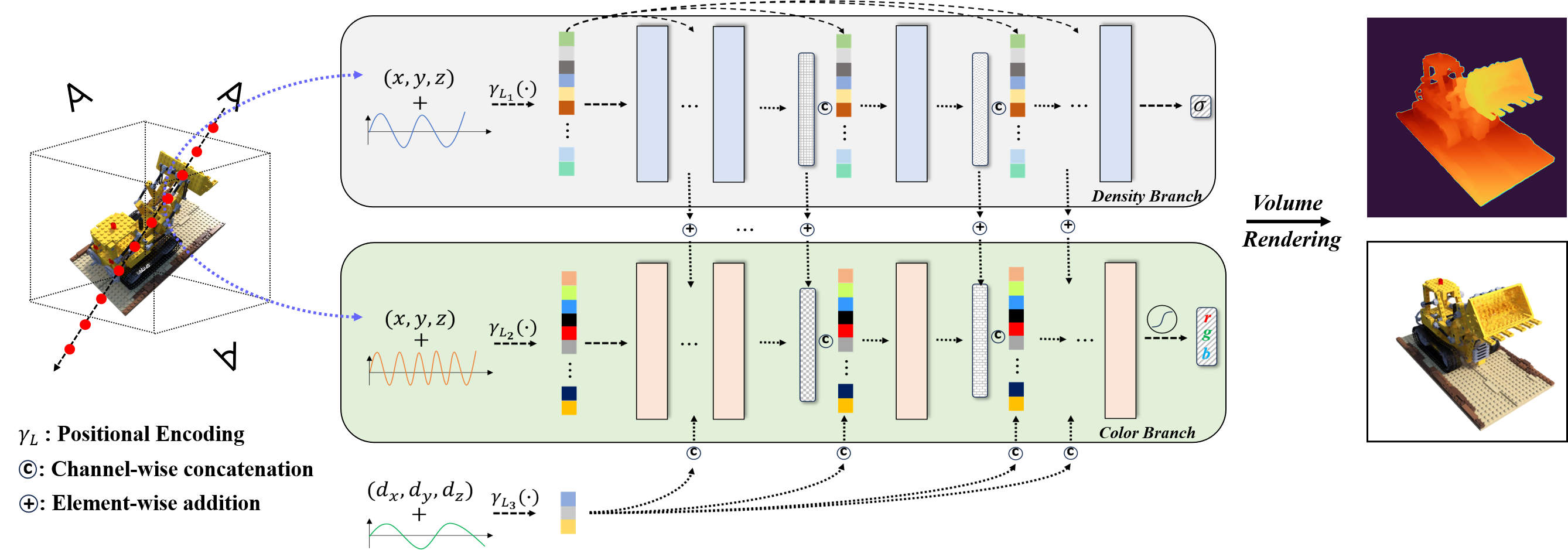}
    \caption{\textbf{Network structure of our proposed method.} To avoid the overfitting issue in few-shot view synthesis, we propose multi-input MLP (mi-MLP) that incorporates inputs (\ie, location $(x,y,z)$ and viewing direction $(d_x,d_y,d_z)$) into each layer of the MLP (Sec.~\ref{sec:Per-layer Inputs Incorporation}). To further improve geometry recovery, we model volume density and colors separately with different frequencies (Sec.~\ref{sec:Modeling Colors and Volume Density Separately}).}
    \label{fig:framework}
\end{figure*}

\subsection{Few-shot View Synthesis}
\paragraph{Prior-based methods.}
Prior-based approaches enable NeRF for few-shot view synthesis either by training a generalized model through large datasets of different scenes or by introducing off-the-shelf pre-trained models. Early works~\cite{yu2021pixelnerf,trevithick2021grf,wang2021ibrnet,chen2021mvsnerf} extracted convolutional features from input views as conditions to render novel views, using classical graphics pipelines such as image-based rendering~\cite{shum2000review,chan2007image} or multi-view stereo~\cite{furukawa2015multi,seitz2006comparison}. VisionNeRF~\cite{lin2023vision}, however, used vision transformers to extract both local and global features for occlusion-aware rendering. DSNeRF~\cite{deng2022depth} and DDP-NeRF~\cite{roessle2022dense} further used depth information obtained from Structure-From-Motion~\cite{schonberger2016structure} or pre-trained depth completion models to incorporate explicit 3D priors. More recently, SparseNeRF~\cite{wang2023sparsenerf} proposed to utilize depth priors obtained from real-world inaccurate observations. DiffusioNeRF~\cite{wynn2023diffusionerf} learned priors over scene geometry and colors through a more powerful diffusion model, which is trained on RGBD patches. While these methods can produce photorealistic novel views, they often require expensive pre-training costs, and the pre-trained scenes may not be suitable for the target scene.

\vspace{-0.4cm}
\paragraph{Regularization-based methods.} 
Regularization-based methods instead obey a per-scene optimization manner similar to vanilla NeRF~\cite{mildenhall2021nerf}, and introduce additional regularization terms or training sources for better novel view synthesis. Specifically, semantic consistency loss~\cite{jain2021putting}, depth-smoothing loss~\cite{niemeyer2022regnerf}, and ray-entropy loss~\cite{kim2022infonerf} were first introduced to constrain unseen views for better geometry recovery. To increase the number of training views available, several works~\cite{ahn2022panerf,chen2022geoaug,kwak2023geconerf,xu2022sinnerf} proposed to use depth-warping to generate novel view images as pseudo labels. Recently, FreeNeRF~\cite{yang2023freenerf} followed a coarse-to-fine manner through a novel frequency annealing strategy on positional encoding. MixNeRF~\cite{seo2023mixnerf} modeled rays as mixtures of Laplacianssians, followed by FlipNeRF~\cite{seo2023flipnerf} which uses flipped reflection rays as additional training sources. SimpleNeRF~\cite{somraj2023simplenerf} proposed to use augmented models to avoid overfitting, which performs well on forward-facing scenes. Though remarkable results have been achieved, all these methods still use the network structure proposed by vanilla NeRF. In contrast, in this paper, we achieve the few-shot view synthesis from the perspective of designing a better network structure.

\section{Preliminaries: NeRF}
\label{sec:preliminaries}
Different from classical explicit scene representation methods such as mesh, voxel, and point cloud, Neural Radiance Field (NeRF)~\cite{mildenhall2021nerf} utilizes an MLP $F_\theta$ to represent scenes implicitly and compactly. For a ray ${\textbf{\textit{r}}}$ cast from camera origin ${\textbf{\textit{o}}}$ through a pixel ${\textbf{\textit{p}}}$ along direction ${\textbf{\textit{d}}}$, a point ${\textbf{\textit{r}}}_t = {\textbf{\textit{o}}} + t{\textbf{\textit{d}}}$ is first sampled from the ray, where $t\in[t_\text{near}, t_\text{far}]$. Subsequently, ${\textbf{\textit{r}}}_t$ is sent to $F_\theta$ to estimate the scene properties, \ie, the corresponding color ${\textbf{\textit{c}}}$ and volume density $\sigma$, which is denoted as:
\begin{equation}
  \centering \label{equ_1}
    {\textbf{\textit{c}}},\sigma = F_{\theta}(\gamma_L({\textbf{\textit{r}}}_t),\gamma_L({\textbf{\textit{d}}})),
\end{equation}
where $\gamma$ is the positional encoding operation aimed at obtaining high-frequency details that is formulated as follows:
\begin{equation}
\centering \label{equ_2}
  \gamma_L(\mathbf{\textit{\textbf{x}}})=(\sin(2^0\mathbf{\textit{\textbf{x}}}),\cos(2^0\mathbf{\textit{\textbf{x}}}),\cdots, \sin(2^{L-1}\mathbf{\textit{\textbf{x}}}),\cos(2^{L-1}\mathbf{\textit{\textbf{x}}})),
\end{equation}
where $L$ is a hyperparameter that controls the frequencies.

Given the color and volume density of $\mathbf{\textbf{\textit{r}}}_t$, the color of ray $\mathbf{\textbf{\textit{r}}}$ can be estimated using the following equation:
\begin{equation}
  \centering \label{equ_3}
    \mathbf{C}(\mathbf{\textbf{\textit{r}}})=\int_{t_{\text{near}}}^{t_{\text{far}}} T(t) \sigma(\mathbf{\textbf{\textit{r}}}(t)) \mathbf{\textbf{\textit{c}}}(\mathbf{\textbf{\textit{r}}}(t), \mathbf{\textbf{\textit{d}}}) d t,
\end{equation}
where $T(t)=\exp \left(-\int_{t_\text{near}}^{t} \sigma(\mathbf{\textbf{\textit{r}}}(s)) d s\right)$ represents the accumulated transmittance. The NeRF is then optimized using common reconstruction loss, \ie, 
\begin{equation}
  \centering \label{equ_4}
    \mathcal{L}=\frac{1}{|\mathcal{R}|} \sum_{\mathbf{\textbf{\textit{r}}} \in \mathcal{R}}\|\mathbf{C}(\mathbf{\textbf{\textit{r}}})-\mathbf{C}_{\text{gt}}\|_2^2,
\end{equation}
where $\mathcal{R}$ is a batch of sampling rays, $\mathbf{C}(\mathbf{\textbf{\textit{r}}})$ is obtained by Eq.~\ref{equ_3} and $\mathbf{C}_{\text{gt}}$ represents the ground-truth color.

\section{Methods}
\paragraph{Motivation.}
As mentioned in Sec.~\ref{sec:intro}, when only a few input views are available, NeRF faces a significant challenge of overfitting. To solve this problem, we drew inspiration from two key observations: 1) as illustrated in FreeNeRF~\cite{yang2023freenerf}, the overfitting issue is caused by the over-fast convergence speed of NeRF on high-frequency details. In this way, the model quickly memorizes input views instead of correctly inferring the underlying geometry. Hence, to avoid overfitting, a direct solution is to decrease the model capacity by reducing the model parameters (\eg, reducing MLP layers); 2) however, though such a simple operation can alleviate overfitting to some extent, as presented in DietNeRF~\cite{jain2021putting}, this simplified NeRF is hardly to recover accurate details, resulting in blurry novel views.

Based on the two observations above, to achieve few-shot view synthesis, our intuition is that in the initial stages of training, the model capacity should be restricted to prevent NeRF from memorizing input views and thus avoid overfitting. However, during the later stage of training, the model capacity should be preserved for detailed rendering.

\subsection{Network Structure}\label{sec:Network structure}

Our network consists of two designs as elaborated in Sec.~\ref{sec:Per-layer Inputs Incorporation} and Sec.~\ref{sec:Modeling Colors and Volume Density Separately} respectively. The resulting architecture is illustrated in Fig.~\ref{fig:framework}.

\subsubsection{Per-layer Inputs Incorporation} \label{sec:Per-layer Inputs Incorporation}
We address the overfitting problem in the few-shot view synthesis from the perspective of network structure. Specifically, as shown in Fig.~\ref{fig: freeze paras}(b), we propose multi-input MLP (mi-MLP) that incorporates inputs (\ie, 3D location and 2D viewing direction) into each layer of the MLP, which is formulated as follows:
\begin{equation}
  \centering \label{equ_5}
  \textbf{\textit{f}}_i = \phi_i(\textbf{\textit{f}}_{i-1},\gamma_L(\textbf{\textit{x}})),\ \ \textbf{\textit{f}}_1 = \phi_1(\gamma_L(\textbf{\textit{x}})),
\end{equation}
where $\phi_i$ is the $i$-th ($i>2$) layer of the MLP, $\textbf{\textit{f}}_i$ is the corresponding output feature, $\textbf{\textit{x}}$ is the input 5D coordinate and $\gamma_L(\textbf{\textit{x}})$ represents the encoded input embeddings (Eq.~\ref{equ_2}).

In contrast to vanilla NeRF, which uses all layers to learn mappings from input embeddings to outputs as shown in Fig.~\ref{fig: freeze paras}(a), our formulation ensures that each layer of the MLP is aware of the input embeddings explicitly. This allows the mappings from input embeddings to outputs with varying number of layers. We hypothesize that such flexible connections between inputs and outputs play a significant role in alleviating the overfitting issue. The analysis is provided below.

\vspace{-0.4cm}
\paragraph{How mi-MLP works?}
\begin{figure}[t]
    \centering
    \begin{subfigure}[b]{1.0\linewidth}
         \centering
    \includegraphics[width=1\linewidth]{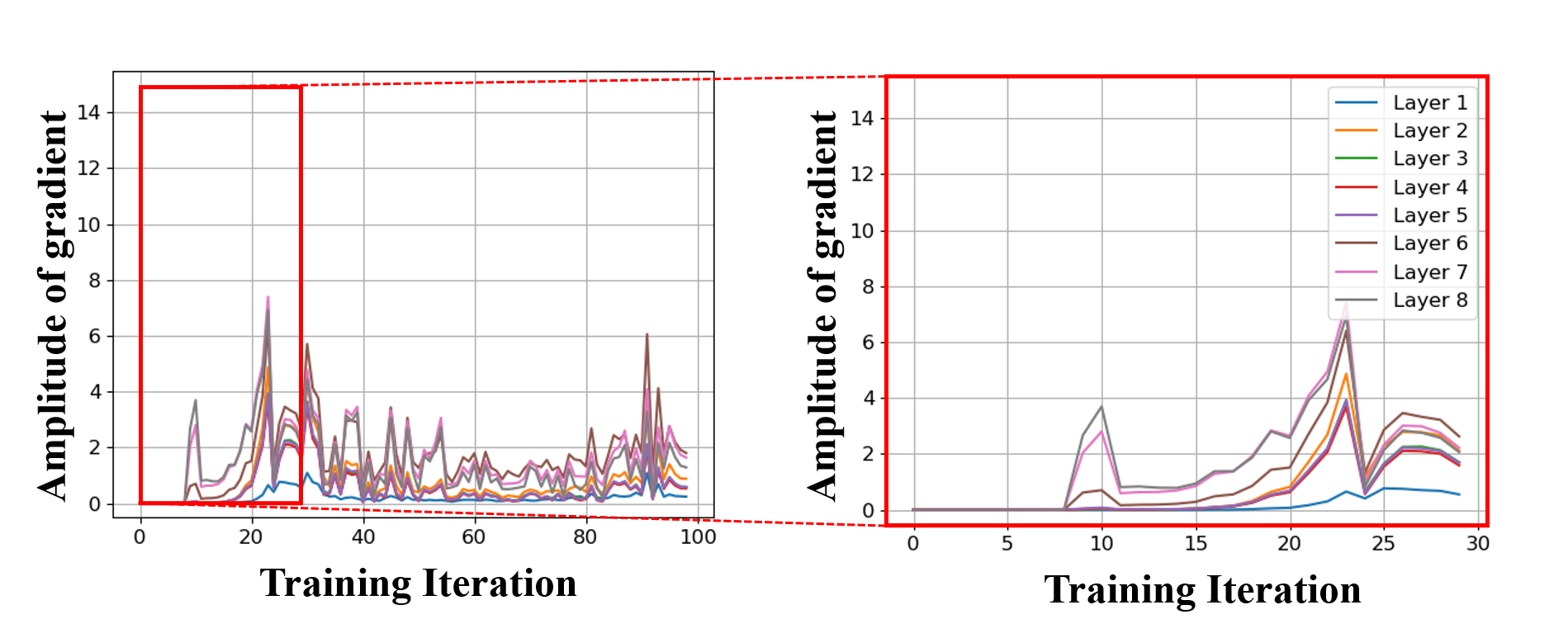}
    \caption{Amplitude of gradients of each layer in vanilla MLP in NeRF.}
    \end{subfigure}
    \begin{subfigure}[b]{1.0\linewidth}
         \centering
    \includegraphics[width=1\linewidth]{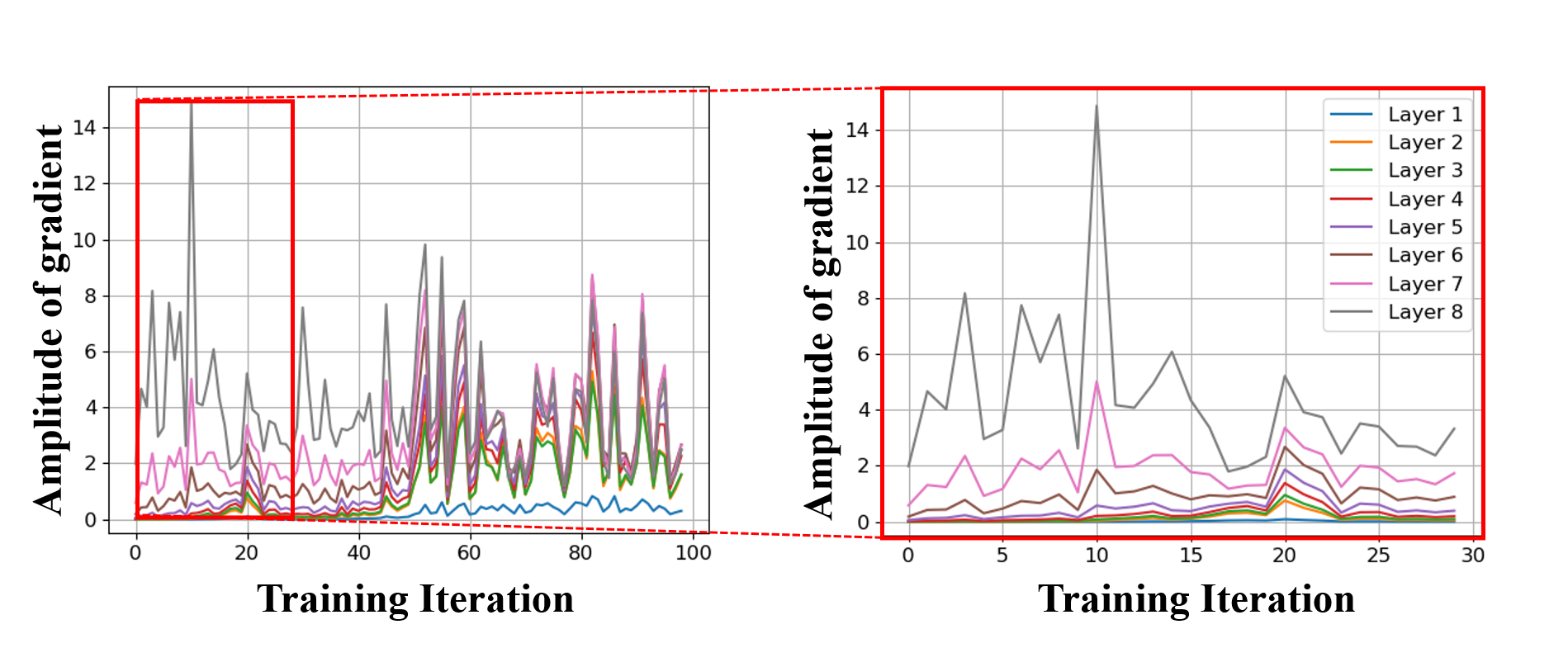}
    \caption{Amplitude of gradients of each layer in our proposed mi-MLP.}
    \end{subfigure}
    \caption{Illustration of the averaged amplitude of gradients of each layer in MLP at the beginning of training. (a) All layers in vanilla MLP have a similar amplitude of gradients. (b) In contrast, mi-MLP enables that the deeper layers (\ie, layers close to the outputs) are updated with large gradients while the shallower layers are updated with extremely small ones.}
    \label{fig:updating_degree}
\end{figure}

Intuitively, the per-layer inputs incorporation enables shorter paths between inputs and outputs, allowing synthesis with fewer parameters in an end-to-end way. It also encourages that the amplitude of gradients of the shallower layer be smaller than that of the deeper layer. As demonstrated in Fig.~\ref{fig:updating_degree}, at the beginning of the training stage, in contrast to vanilla MLP that results in a similar amplitude of gradients for each layer (Fig.~\ref{fig:updating_degree}(a)), mi-MLP enables that the deeper layers (\ie, layers close to the outputs) are updated with large gradients while the shallower layers are updated with extremely small ones (Fig.~\ref{fig:updating_degree}(b)). 


More theoretical, assuming $\gamma_L(\textbf{\textit{x}})\in\mathbb{R}^{d_1 \times 1}$, $\textbf{\textit{f}}_i\in\mathbb{R}^{d_2 \times 1}$, the bias vector and weight matrix of $\phi_i$ are $\textbf{\textit{b}}_i\in\mathbb{R}^{d_2 \times 1}$ and $\textbf{\textit{w}}_i=(\textbf{\textit{w}}_i^1,\textbf{\textit{w}}_i^2,\dots,\textbf{\textit{w}}_i^{d_2})^T$ respectively, where $\textbf{\textit{w}}_i^j=(\textbf{\textit{w}}_i^{j0}\in\mathbb{R}^{1 \times d_1},\textbf{\textit{w}}_i^{j1}\in\mathbb{R}^{1 \times d_2})^T$. Thus Eq.~\ref{equ_5} is equivalent to 
\begin{equation}
\begin{split}
  \centering \label{equ_6}
  \phi_i^j(\gamma_L(\textbf{\textit{x}})) &= \epsilon \{\textbf{\textit{w}}_i^j[\gamma_L(\textbf{\textit{x}}),\phi_{i-1}(\gamma_L(\textbf{\textit{x}}))]^T+\textbf{\textit{b}}_i\} \\
  &= \epsilon \{\textbf{\textit{w}}_i^{j0}[\gamma_L(\textbf{\textit{x}})]+\textbf{\textit{w}}_i^{j1}[\phi_{i-1}(\gamma_L(\textbf{\textit{x}}))]+\textbf{\textit{b}}_i\},
\end{split}
\end{equation}
where $\phi_i^j$ is the $j$-th element of $\textbf{\textit{f}}_i$, $\epsilon$ denotes the activation function whose default setting is ReLU. It can be proved that the closed-form solution that represents the ratio of the amplitude of gradients of two adjacent layers can be formulated as follows, where $\mathcal{L}$ means the loss function:
\begin{equation}
\begin{split}
\centering \label{equ_7}
  &\Vert \frac{\partial \mathcal{L}}{\partial \textbf{\textit{w}}_i} \Vert_1 / \Vert \frac{\partial \mathcal{L}}{\partial \textbf{\textit{w}}_{i-1}} \Vert_1 = \frac{1}{d_2}\sum_{j=1}^{d_2}\Vert \frac{\partial \mathcal{L}}{\partial \textbf{\textit{w}}_i^j} \Vert_1 / \Vert \frac{\partial \mathcal{L}}{\partial \textbf{\textit{w}}_{i-1}^j} \Vert_1 \\
  =& \frac{1}{d_2}\sum_{j=1}^{d_2} \frac{\Vert \gamma_L(\textbf{\textit{x}}) \Vert_1 + \Vert \phi_{i-1}(\gamma_L(\textbf{\textit{x}})) \Vert_1}{\Vert \sum\textbf{\textit{w}}_i^{j1}\Vert_1\cdot\{\Vert \gamma_L(\textbf{\textit{x}}) \Vert_1 + \Vert \phi_{i-2}(\gamma_L(\textbf{\textit{x}})) \Vert_1\}},
\end{split}
\end{equation}
Accordingly, $\Vert \frac{\partial \mathcal{L}}{\partial \textbf{\textit{w}}_i} \Vert_1 / \Vert \frac{\partial \mathcal{L}}{\partial \textbf{\textit{w}}_{i-1}} \Vert_1\ge 1$ holds true in a high probability during the early stage of training when $\Vert \sum\textbf{\textit{w}}_i^{j1}\Vert_1\in(0,1]$ and $ \Vert \phi_{i-1}(\gamma_L(\textbf{\textit{x}})) \Vert_1 \approx \Vert \sum\textbf{\textit{w}}_i^{j1}\Vert_1 \cdot \Vert \phi_{i-2}(\gamma_L(\textbf{\textit{x}})) \Vert_1$. In practice, we find that the default initialization provided by PyTorch can meet the requirements, where the amplitude of gradients of each layer is shown in Fig.~\ref{fig:updating_degree}. 
Please refer to the supplementary materials for more details.

\subsubsection{Modeling Colors and Volume Density Separately}\label{sec:Modeling Colors and Volume Density Separately}
Although mi-MLP alone can perform comparably to several prior methods, the rendered novel views still contain noticeable artifacts, as shown in Fig.~\ref{fig:ablation_blender_8}. To address this issue and further improve geometry recovery, we propose to model volume density and colors separately.


Specifically, it is widely accepted that geometry (represented by the volume density) is not as detailed as appearance (represented by the colors), since geometry is usually piecewise smooth~\cite{niemeyer2022regnerf}. To prioritize low-frequency information in volume density, we propose to reduce the dimensions of input embeddings for volume density in comparison to those for colors, considering that the dimensions of the encoded input embeddings obtained by Eq.~\ref{equ_2} decide how detailed the output is~\cite{mildenhall2021nerf,yang2023freenerf}.

To this end, different from NeRF which uses one shared MLP to predict colors and volume density synchronously, we instead use two separate MLPs to estimate them individually, dubbed the Color Branch $C_\theta$ and Density Branch $D_\theta$, where the dimensions of input embeddings for different branches are not the same.
As shown in Fig.~\ref{fig:framework}, the whole network structure can thus be formulated as follows:
\begin{equation}
  \centering \label{equ_8}
  \sigma = D_\theta(\gamma_{L_1}(\textbf{\textit{x}})), \textbf{\textit{c}} = C_\theta(\gamma_{L_2}(\textbf{\textit{x}}), \gamma_{L_3}(\textbf{\textit{d}})),
\end{equation}
where $\sigma$ and $\textbf{\textit{c}}$ denote the estimated volume density and colors respectively,  $\textbf{\textit{x}}$ is the input 3D point coordinate, $\textbf{\textit{d}}$ is viewing direction vector, $L_1, L_2$, and $L_3$ are hyperparameters that control the frequencies of positional encoding which satisfy $L_3 \leq L_1 \leq L_2$. 

Overall, we adopt both per-layer incorporation and separate modeling of colors and volume density in our network design.
Therefore, for the Density Branch, as illustrated in Sec.~\ref{sec:Per-layer Inputs Incorporation}, we incorporate inputs into each layer, \ie,
\begin{equation}
  \centering \label{equ_9}
  \textbf{\textit{f}}_i^D = \phi_i^D(\textbf{\textit{f}}_{i-1}^D,\gamma_{L_1}(\textbf{\textit{x}})),\ \ \textbf{\textit{f}}_1^D = \phi_1^D(\gamma_{L_1}(\textbf{\textit{x}})),
\end{equation}
where $\phi_i^D$ is the $i$-th ($i \ge 2$) layer of the Density Branch MLP, $\textbf{\textit{f}}_i^D$ is the corresponding output feature. For the Color Branch, we empirically find that an interaction between the Color Branch and the Density Branch is beneficial to better geometry recovery, which is denoted as follows:
\begin{equation}
\begin{split}
  \centering \label{equ_10}
  &\textbf{\textit{f}}_{i-1}^C = \phi_{i-1}^C(\textbf{\textit{f}}_{i-2}^C, \gamma_{L_3}(\textbf{\textit{d}})) + \textbf{\textit{f}}_{i-1}^D \\ &\textbf{\textit{f}}_i^C = \phi_i^C(\textbf{\textit{f}}_{i-1}^C, \gamma_{L_3}(\textbf{\textit{d}})),\ \
  \textbf{\textit{f}}_1^C = \phi_1^C(\gamma_{L_2}(\textbf{\textit{x}})),
\end{split}
\end{equation}
where $\phi_i^C$ is the $i$-th ($i \ge 2$) layer of the Color Branch MLP, $\textbf{\textit{f}}_i^C$ is the corresponding output feature.

\begin{figure}[t]
    \centering
    \includegraphics[width=1\linewidth]{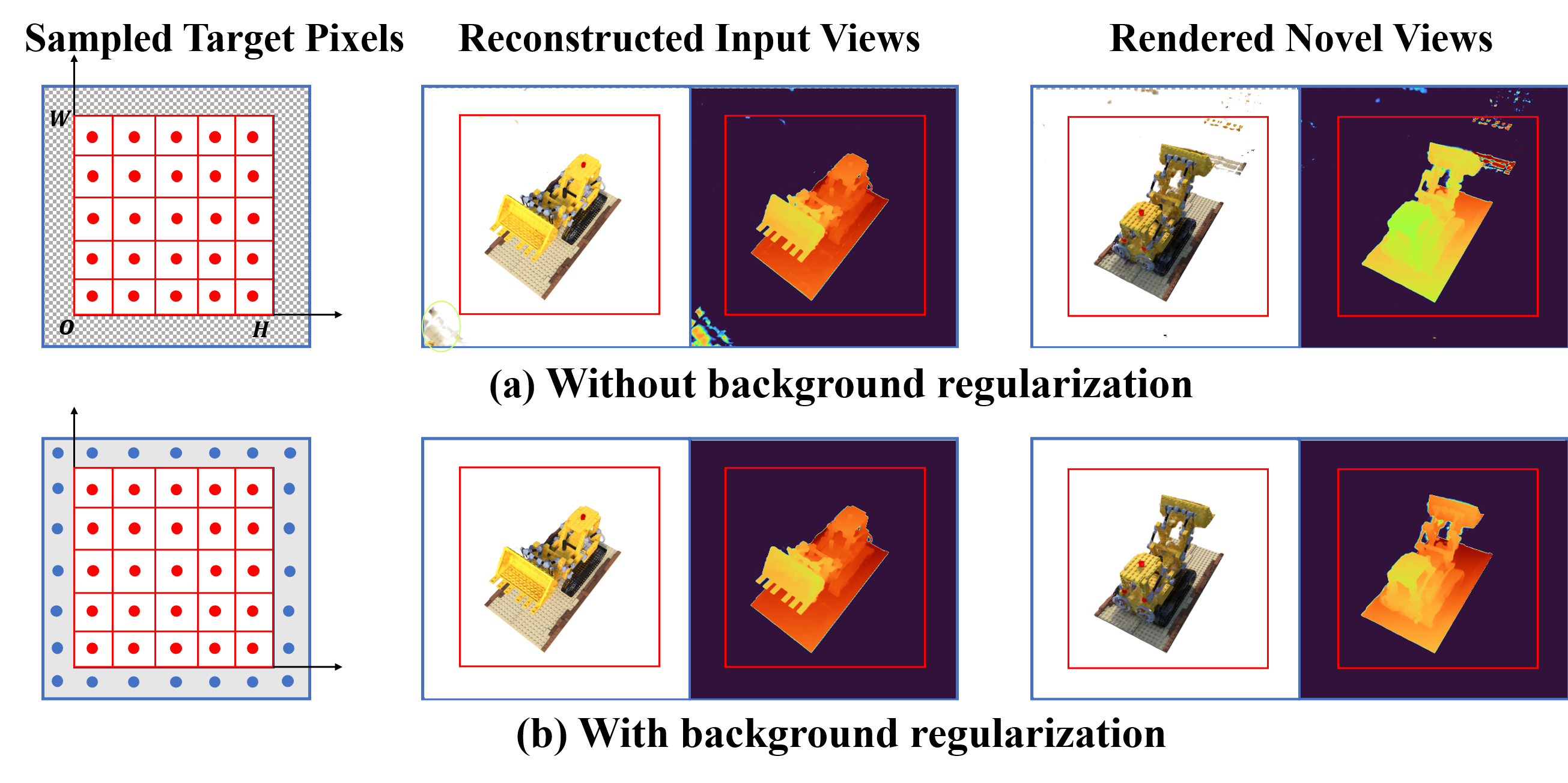} 
    \vspace{-6mm}
    \caption{\textbf{Background regularization.} In addition to sampling target pixels within the image space (\ie, the red dots) to generate training rays, we also sample target pixels outside the image space (\ie, the blue dots) to address background artifacts in object-centric scenes.}
    \label{fig:background regularization}
\end{figure}

\begin{figure}[t]
    \centering
    \includegraphics[width=1\linewidth]{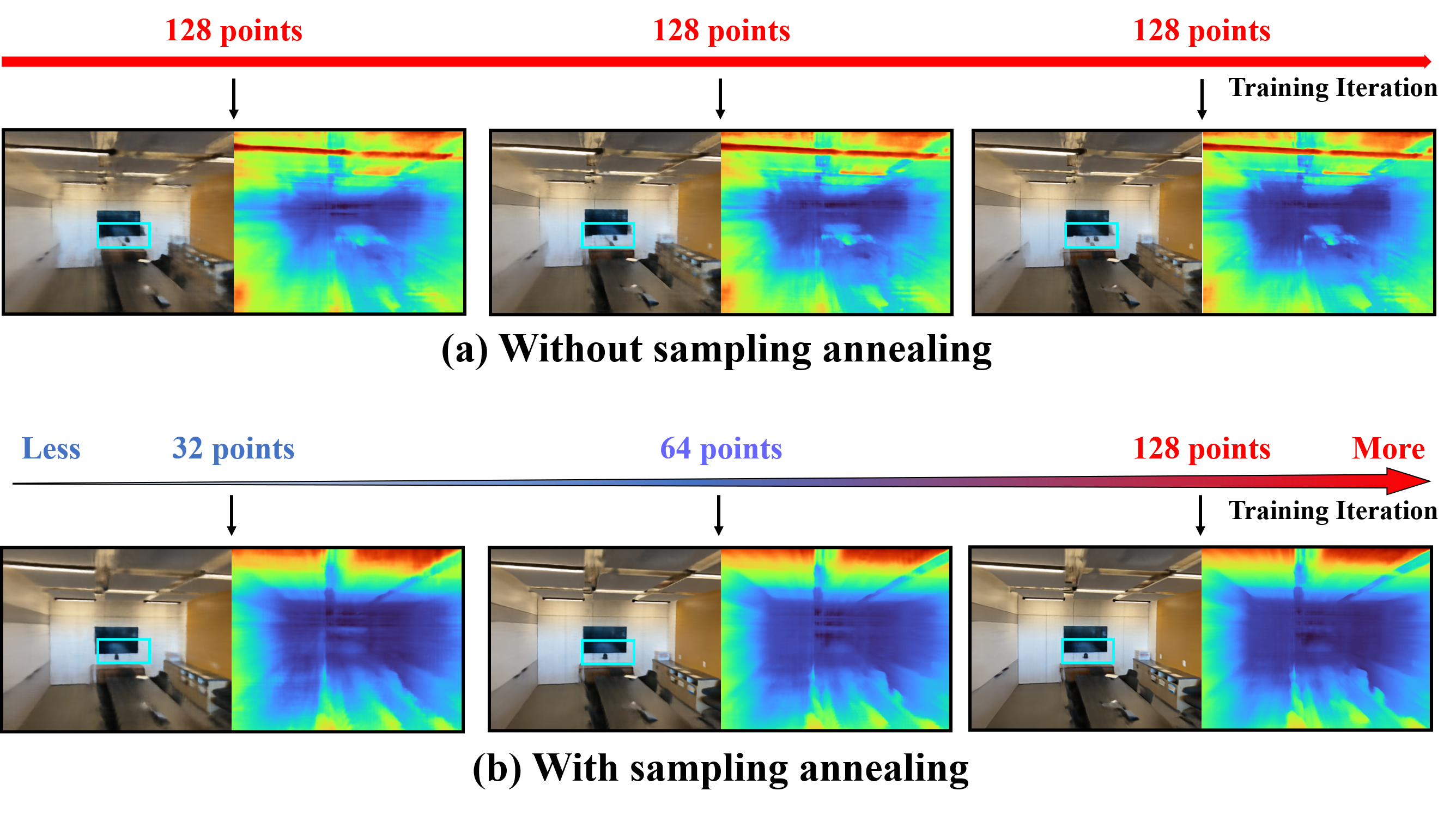}
    \vspace{-6mm}
    \caption{\textbf{Sampling annealing.} During the early stage of training, fewer points are sampled along a ray to make the network more focused on coarse geometry estimation, while more sampling points are utilized during the later stage for details recovery.}
    \vspace{-3mm}
    \label{fig: Sampling Annealing}
\end{figure}

\begin{figure*}
    \centering
    \begin{tabular}{P{0.2\textwidth}P{0.2\textwidth}P{0.2\textwidth}P{0.18\textwidth}P{0.07\textwidth}}
     \scriptsize DietNeRF~\cite{jain2021putting} &  \scriptsize InfoNeRF~\cite{kim2022infonerf} & \scriptsize FreeNeRF~\cite{yang2023freenerf} & \scriptsize Ours & \scriptsize GT\\
    \end{tabular}
    \begin{subfigure}[b]{1.0\linewidth}
         \centering 
    \includegraphics[width=\linewidth]
    {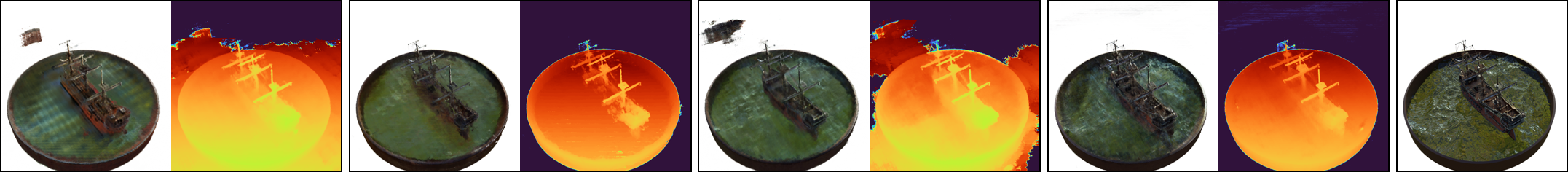}
    \caption{View synthesis and estimated depth map on Blender with 8 input views.}
    \end{subfigure}
    \begin{tabular}{P{0.2\textwidth}P{0.2\textwidth}P{0.2\textwidth}P{0.18\textwidth}P{0.07\textwidth}}
     \scriptsize MVSNeRF-ft~\cite{chen2021mvsnerf} &  \scriptsize RegNeRF~\cite{niemeyer2022regnerf} & \scriptsize FreeNeRF~\cite{yang2023freenerf} & \scriptsize Ours & \scriptsize GT\\
     \end{tabular}
    \begin{subfigure}[b]{1.0\linewidth}
         \centering 
    \includegraphics[width=\linewidth]
    {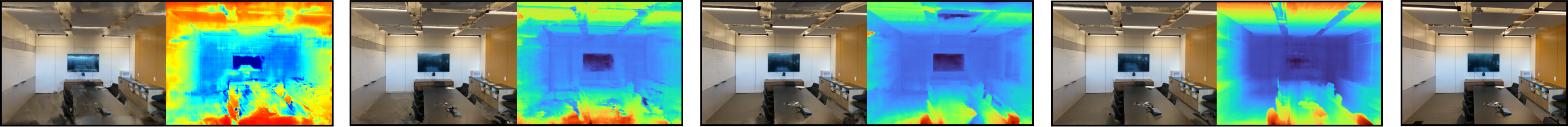}
    \caption{View synthesis and estimated depth map on LLFF with 3 input views.}
    \end{subfigure}
    \begin{tabular}{P{0.2\textwidth}P{0.2\textwidth}P{0.2\textwidth}P{0.18\textwidth}P{0.07\textwidth}}
     \scriptsize MVSNeRF-ft~\cite{chen2021mvsnerf} &  \scriptsize RegNeRF~\cite{niemeyer2022regnerf} & \scriptsize FreeNeRF~\cite{yang2023freenerf} & \scriptsize Ours & \scriptsize GT\\
     \end{tabular} 
    \begin{subfigure}[b]{1.0\linewidth}
         \centering 
    \includegraphics[width=\linewidth]
    {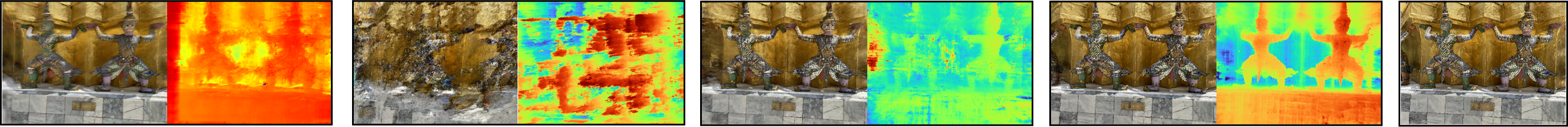}
    \caption{View synthesis and estimated depth map on Shiny with 3 input views.}
    \end{subfigure}
    \vspace{-6mm}
\caption{
    \textbf{Qualitive comparisons on the Blender, LLFF, and Shiny dataset.} Our proposed method can achieve both photorealistic novel views and accurate depth estimation, ft indicates the results fine-tuned on each scene individually.
    }
    \vspace{-3mm}
    \label{fig:Qualitive comparisons}
\end{figure*}

\subsection{Background Regularization}
A common failure mode for rendering scenes centered on a single object is the presence of background artifacts for both reconstructed input views and rendered novel views, as shown in Fig.~\ref{fig:background regularization}(a).

We assume that this is caused by insufficient constraints on the background. Specifically, during the training process of NeRF, the sampled target pixels $\textbf{\textit{p}} = (p_x,p_y)$ that generate training rays $\textbf{\textit{r}}$ are all distributed inside the input image space, where $p_x\in[0, H], p_y\in[0, W]$, $H$ and $W$ represent the height and width of input images. For object-centric scenes, it is reasonable to assume that the corresponding pixel colors outside the image space should be the same as the background color. However, as shown in Fig.~\ref{fig:background regularization}(a), when only a few input views are available, the extrapolated input image contains apparent artifacts, especially in the areas that lie outside the image space.

Motivated by this observation, we propose a regularization technique for background artifact removal, which is denoted as follows:
\begin{equation}
  \centering \label{equ_11}
  \mathcal{L}_{\text{BR}}=\frac{1}{|\mathcal{R}_o|} \sum_{\mathbf{\textbf{\textit{r}}} \in \mathcal{R}_o}\|\mathbf{C}(\mathbf{\textbf{\textit{r}}})-\mathbf{C}_{\text{bk}}\|_2^2,
\end{equation}
where $\mathcal{R}_o$ is a batch of sampling rays generated from target pixels outside the input image space, $\mathbf{C}(\mathbf{\textbf{\textit{r}}})$ is the rendered color and $\mathbf{C}_{\text{bk}}$ is the background color. As shown in Fig.~\ref{fig:background regularization}(b), the regularization obtains clear images effectively.

\subsection{Sampling Annealing}
In the context of real-world scenes, as shown in Fig.~\ref{fig: Sampling Annealing}(a), we also observe that the floating artifacts are distributed in close proximity to the camera~\cite{niemeyer2022regnerf,yang2023freenerf}, which are referred to near-field artifacts. 

To solve this problem, we propose a sampling annealing strategy, where the number of sampling points along a ray increases linearly during training, which is formulated as follows:
\begin{equation}
  \centering \label{equ_12}
  N_t = \min(N_{\max}, \lfloor u / \eta \rfloor + N_{start}),
\end{equation}
where $u$ denotes the current training iteration, $N_t$ indicates the number of sampling points along one ray at the $u$-th iteration, $N_{\max}$ is the maximum number of sampling points, $N_{start}$ is the number of sampling points at the start of training, $\eta$ is a hyperparameter that controls the increasing speed of sampling points.

\section{Experiments}
\vspace{-0.2cm}
\paragraph{Datasets and metrics.}
We evaluate our proposed method on three popular datasets: Blender~\cite{mildenhall2021nerf}, LLFF~\cite{mildenhall2019local}, and Shiny~\cite{wizadwongsa2021nex}. Blender consists of 8 synthetic 360$^{\circ}$ object-centric scenes with white background. LLFF and Shiny individually contain 8 real-world forward-facing scenes, while Shiny is much more complex due to its view-dependent effects, such as reflections and refraction. We follow the experimental protocols provided by~\cite{niemeyer2022regnerf,jain2021putting}.

We use PSNR, SSIM~\cite{wang2004image}, and LPIPS~\cite{zhang2018unreasonable} to measure the quantitative results of our proposed methods. We also report the geometric average following~\cite{niemeyer2022regnerf} for an easier comparison. See more experimental details in the supplementary materials.

\begin{table*}
\fontsize{6}{6}\selectfont
    \resizebox{\linewidth}{!}{\begin{tabular}{l|c|cc|cc|cc|cc}
    \toprule
          \multirow{2}{*}{Method} & \multirow{2}{*}{Setting} & \multicolumn{2}{c}{PSNR$\uparrow$} & \multicolumn{2}{c}{SSIM$\uparrow$} & \multicolumn{2}{c}{LPIPS$\downarrow$} & \multicolumn{2}{c}{Average$\downarrow$} \\
          &       & 4-view & 8-view & 4-view & 8-view & 4-view & 8-view & 4-view & 8-view \\
    \midrule
    \midrule
    Vanilla-NeRF~\cite{mildenhall2021nerf}  & \multirow{3}{*}{Baselines}   & -     & \cellcolor{gray!25}14.73     & -     & \cellcolor{gray!25}0.734     & -     &  \cellcolor{gray!25}0.451     & -     & \cellcolor{gray!25}0.199 \\
    Mip-NeRF~\cite{barron2021mip}  &    & 14.12     & 18.74     & 0.722     &   0.828    & 0.382     &  0.238      & 0.221     & 0.121 \\
    Ref-NeRF~\cite{verbin2022ref}  &    & 18.09     & \cellcolor{yellow!25}24.00     & 0.764     & \cellcolor{yellow!25}0.879     & 0.269     &  0.106     & 0.150     & \cellcolor{yellow!25}0.059 \\
    \midrule
    NV~\cite{lombardi2019neural} & \multirow{7}{*}{Regularization-based}   & -     & 17.85     & -     & 0.741     & -     &  0.245    & -     & 0.127 \\
    Simplified NeRF~\cite{jain2021putting} &   & -     & 20.09     & -     & 0.822     & -     &  0.179    & -     & 0.090 \\
    DietNeRF~\cite{jain2021putting} &   & 15.42     & 23.14    & 0.730     & 0.866     & 0.314     &  0.109   & 0.201     & 0.063 \\
    InfoNeRF~\cite{kim2022infonerf} &   & 18.44     & 22.01     & 0.792     & 0.852     & 0.223     &  0.133    & 0.119     & 0.073 \\
    RegNeRF~\cite{niemeyer2022regnerf} &   & 13.71     & 19.11     & 0.786     & 0.841     & 0.346     &  0.200    & 0.210     & 0.122 \\
    MixNeRF~\cite{seo2023mixnerf} &   & \cellcolor{yellow!25}18.99     & 23.84     & \cellcolor{yellow!25}0.807     & 0.878     & \cellcolor{yellow!25}0.199     &  \cellcolor{yellow!25}0.103    & \cellcolor{yellow!25}0.113     & 0.060 \\
    FreeNeRF~\cite{yang2023freenerf} &   & \cellcolor{orange!25}19.70     & \cellcolor{orange!25}24.26     & \cellcolor{orange!25}0.812     & \cellcolor{orange!25}0.883     & \cellcolor{orange!25}0.175     &  \cellcolor{orange!25}0.098    & \cellcolor{orange!25}0.093     & \cellcolor{orange!25}0.058 \\
    \midrule
    \textbf{Ours} & \multirow{1}{*}{Network-based} & \cellcolor{red!25}20.38     & \cellcolor{red!25}24.70     & \cellcolor{red!25}0.828     & \cellcolor{red!25}0.885     & \cellcolor{red!25}0.156     &  \cellcolor{red!25}0.087    & \cellcolor{red!25}0.084     & \cellcolor{red!25}0.047 \\
    \bottomrule
\end{tabular}%
}
    \vspace{-2mm}
    \caption{
    \textbf{Quantitative Comparison on Blender.} Our proposed method can achieve state-of-the-art performance on all metrics. The best, second-best, and third-best entries are marked in red, orange, and yellow, respectively. Our baseline is marked in gray.
    }
    \label{tab:blender}
\end{table*}

\begin{table*}
    \resizebox{\linewidth}{!}{\begin{tabular}{l|c|ccc|ccc|ccc|ccc}
\toprule
  \multirow{2}{*}{Method}& \multirow{2}{*}{Setting} &  \multicolumn{3}{c}{PSNR $\uparrow$} & \multicolumn{3}{c}{SSIM $\uparrow$} & \multicolumn{3}{c}{LPIPS $\downarrow$} & \multicolumn{3}{c}{Average $\downarrow$}  \\
  &  & 3-view & 6-view & 9-view  & 3-view & 6-view & 9-view  & 3-view & 6-view & 9-view  & 3-view & 6-view & 9-view \\ \midrule
\midrule  
Vanilla-NeRF~\cite{mildenhall2021nerf} & \multirow{2}{*}{Baselines} & \cellcolor{gray!25}13.34 & - & - & \cellcolor{gray!25}0.373 & - & - & \cellcolor{gray!25}0.451 & - & - & \cellcolor{gray!25}0.255 & - & - \\
Mip-NeRF~\cite{barron2021mip} &  & 14.62 & 20.87 & 24.26 & 0.351 & 0.692 & 0.805 & 0.495 & 0.255 & 0.172 & 0.246 & 0.114 & 0.073 \\
\midrule
SRF~\cite{chibane2021stereo} & \multirow{6}{*}{Prior-based} & 12.34 & 13.10 & 13.00 & 0.250 & 0.293 & 0.297 & 0.591 & 0.594 & 0.605 & 0.313 & 0.293 & 0.296 \\
PixelNeRF~\cite{yu2021pixelnerf} &  & 7.93 & 8.74 & 8.61 & 0.272 & 0.280 & 0.274 & 0.682 & 0.676 & 0.665 & 0.461 & 0.433 & 0.432 \\
MVSNeRF~\cite{chen2021mvsnerf} &  & 17.25 & 19.79 & 20.47 & 0.557 & 0.656 & 0.689 & 0.356 & 0.269 & 0.242 & 0.171 & 0.125 & 0.111 \\
SRF-ft~\cite{chibane2021stereo} &  & 17.07 & 16.75 & 17.39 & 0.436 & 0.438 & 0.465 & 0.529 & 0.521 & 0.503 & 0.203 & 0.207 & 0.193 \\
PixelNeRF-ft~\cite{yu2021pixelnerf} &  & 16.17 & 17.03 & 18.92 & 0.438 & 0.473 & 0.535 & 0.512 & 0.477 & 0.430 & 0.217 & 0.196 & 0.163 \\
MVSNeRF-ft~\cite{chen2021mvsnerf} &  & 17.88 & 19.99 & 20.47 & 0.584 & 0.660 & 0.695 & \cellcolor{yellow!25}0.327 & 0.264 & 0.244 & 0.157 & 0.122 & 0.111 \\
\midrule 
DietNeRF~\cite{jain2021putting} & \multirow{3}{*}{\shortstack{Regularization- \\ based}} & 14.94 & 21.75 & 24.28 & 0.370 & 0.717 & 0.801 & 0.496 & 0.248 & 0.183 & 0.240 & 0.105 & 0.073 \\
RegNeRF~\cite{niemeyer2022regnerf} &  & \cellcolor{yellow!25}19.08 & \cellcolor{yellow!25}23.10 & \cellcolor{yellow!25}24.86 & \cellcolor{yellow!25}0.587 & \cellcolor{yellow!25}0.760 & \cellcolor{yellow!25}0.820 & 0.336 & \cellcolor{yellow!25}0.206 & \cellcolor{yellow!25}0.161 & \cellcolor{yellow!25}0.146 & \cellcolor{yellow!25}0.086 & \cellcolor{yellow!25}0.067 \\
FreeNeRF~\cite{yang2023freenerf} &  & \cellcolor{orange!25}19.63 & \cellcolor{red!25}23.73 & \cellcolor{orange!25}25.13 & \cellcolor{orange!25}0.612 & \cellcolor{orange!25}0.779 & \cellcolor{orange!25}0.827 & \cellcolor{orange!25}0.308 & \cellcolor{orange!25}0.195 & \cellcolor{orange!25}0.160 & \cellcolor{orange!25}0.134 & \cellcolor{orange!25}0.075 & \cellcolor{orange!25}0.064 \\
\midrule 
\textbf{Ours} & \multirow{1}{*}{\shortstack{Network-based}} & \cellcolor{red!25}19.75 & \cellcolor{orange!25}23.57 & \cellcolor{red!25}25.15 & \cellcolor{red!25}0.614 & \cellcolor{red!25}0.788 & \cellcolor{red!25}0.834 & \cellcolor{red!25}0.300 & \cellcolor{red!25}0.163 & \cellcolor{red!25}0.140 & \cellcolor{red!25}0.125 & \cellcolor{red!25}0.069 & \cellcolor{red!25}0.055 \\
\bottomrule
\end{tabular}
}
    \vspace{-2mm}
    \caption{
    \textbf{Quantitative Comparison on LLFF.} Our proposed method outperforms other methods on real-world forward-facing scenes, ft indicates the results fine-tuned on each scene individually.
    }
    \vspace{-2mm}
    \label{tab:llff}
\end{table*}

\begin{table}
\fontsize{7}{8}\selectfont
    \resizebox{\linewidth}{!}{\begin{tabular}{l|c|c|c|c}
    \toprule
          Method & PSNR$\uparrow$  & SSIM$\uparrow$  & LPIPS$\downarrow$ & Average$\downarrow$ \\
    \midrule
    \midrule
    Vanilla-NeRF~\cite{mildenhall2021nerf}   & \cellcolor{gray!25}14.37 & \cellcolor{gray!25}0.309 & \cellcolor{gray!25}0.610  & \cellcolor{gray!25}0.264 \\
    \midrule
    MVSNeRF~\cite{chen2021mvsnerf}   & 16.45  & 0.375 &  0.506 & 0.208 \\
    MVSNeRF-ft~\cite{chen2021mvsnerf}  & \cellcolor{yellow!25}17.08 & \cellcolor{yellow!25}0.408 & \cellcolor{yellow!25}0.475 &  \cellcolor{yellow!25}0.192   \\
    \midrule
    DietNeRF~\cite{jain2021putting}   & 13.12 & 0.341 & 0.646  & 0.295 \\
    InfoNeRF~\cite{kim2022infonerf}   & 12.86 & 0.332 & 0.681  & 0.307 \\
    RegNeRF~\cite{niemeyer2022regnerf}  & 12.76 & 0.287 & 0.621 &  0.302  \\
    FreeNeRF~\cite{mildenhall2021nerf}  & \cellcolor{orange!25}17.20 & \cellcolor{orange!25}0.411 & \cellcolor{orange!25}0.454 & \cellcolor{orange!25}0.187    \\
    \midrule
    \textbf{Ours}  & \cellcolor{red!25}18.25 & \cellcolor{red!25}0.475 & \cellcolor{red!25}0.416  & \cellcolor{red!25}0.165 \\
    \bottomrule
\end{tabular}}
    \vspace{-2mm}
    \caption{
    \textbf{Quantitative Comparison on Shiny.} On the more challenging scenes with complex view-dependent effects such as reflection, our proposed method can still obtain a significant performance improvement when only 3 input views are available, ft indicates the results fine-tuned on each scene individually.
    }
    \vspace{-2mm}
    \label{tab:shiny}
\end{table}

\subsection{Comparison with State-of-the-art Methods}
\vspace{-0.1cm}
\paragraph{Blender.}
Our proposed method achieves state-of-the-art performance on the Blender dataset for both 4 and 8 input views, as shown in Tab.~\ref{tab:blender} and Fig.~\ref{fig:Qualitive comparisons}(a). Notably, for methods such as~\cite{jain2021putting} and~\cite{kim2022infonerf} that impose additional regularizations on unseen views, though reasonable results can be obtained, the rendered novel views include unexpected imaginary contents. For FreeNeRF~\cite{yang2023freenerf}, since the regularization is only applied to known input views, the estimated geometry contain severe floating artifacts, as demonstrated from the death map in Fig.~\ref{fig:Qualitive comparisons}(a). In contrast, our proposed method can achieve photorealistic novel view synthesis as well as clear geometry estimation.

\vspace{-0.1cm}
\paragraph{LLFF.}
We also perform experiments on the LLFF dataset with 3/6/9 known input views. As shown in Tab.~\ref{tab:llff} and Fig.~\ref{fig:Qualitive comparisons}(b), our method generally outperforms other baselines across all settings. For prior-based methods, severe artifacts will be generated due to the domain gap between the training dataset and the testing set. Compared to regularization-based methods, ours can achieve the best performance, except for the PSNR metric when 6 input views are available. We believe this is caused by the choice of different baselines, where we use vanilla NeRF as our baseline, while methods like RegNeRF~\cite{niemeyer2022regnerf} and FreeNeRF~\cite{yang2023freenerf} choose a more powerful baseline, \ie, MipNeRF~\cite{barron2021mip}.

\vspace{-0.1cm}
\paragraph{Shiny.}
On account that the Shiny dataset contains more complex view-dependent effects such as reflection and refraction, most regularization-based methods such as~\cite{niemeyer2022regnerf,jain2021putting} perform even worse than vanilla NeRF, due to the mismatch between introduced regularization terms and actual physical prior, as shown in Tab.~\ref{tab:shiny} and Fig.~\ref{fig:Qualitive comparisons}(c).
Though FreeNeRF~\cite{yang2023freenerf} can still work and produce reasonable results, the rendered novel views contain obvious artifacts. In contrast, our proposed method can achieve a significant performance improvement, both quantitatively and qualitatively. More additional results on the three datasets are provided in the supplementary materials.

\begin{table}
    \resizebox{\linewidth}{!}{\begin{tabular}{c|cccc|cccc}
    \toprule
    \multicolumn{9}{c}{\textbf{Blender}} \\
    \midrule
    \multicolumn{1}{r}{} & \textbf{Pli} & \textbf{Sep} & \textbf{Bkr} & \textbf{Sa} & {PSNR$\uparrow$} & {SSIM$\uparrow$} & {LPIPS$\downarrow$} & {Average$\downarrow$} \\
    \midrule
    {NeRF} & \ding{55}     & \ding{55}     & \ding{55}     & \ding{55}     & 14.73 & 0.734 & 0.451 & 0.199 \\
    \midrule
    \multirow{4}[2]{*}{\textbf{Ours}} & \ding{51}      & \ding{55}      & \ding{55}     & \ding{55}     & 24.12 & 0.879 & 0.114 & 0.053 \\
          & \ding{51}      & \ding{55}     & \ding{51}      & \ding{55}     & \cellcolor{orange!25}24.42 & \cellcolor{orange!25}0.882 & \cellcolor{orange!25}0.092 & \cellcolor{orange!25}0.048 \\
          & \ding{51}      & \ding{51}      & \ding{55}     & \ding{55}     & \cellcolor{yellow!25}24.23 & \cellcolor{yellow!25}0.881 & \cellcolor{yellow!25}0.108 & \cellcolor{yellow!25}0.052 \\
          & \ding{51}      & \ding{51}      & \ding{51}      & \ding{55}     & \cellcolor{red!25}24.70  & \cellcolor{red!25}0.885 & \cellcolor{red!25}0.087 & \cellcolor{red!25}0.046 \\
    \midrule
    \multicolumn{9}{c}{\textbf{LLFF}} \\
    \midrule
    {NeRF} & \ding{55}     & \ding{55}     & \ding{55}     & \ding{55}     & 13.34 & 0.373 & 0.451 & 0.255 \\
    \midrule
    \multirow{4}[2]{*}{\textbf{Ours}} & \ding{51}      & \ding{55}     & \ding{55}     & \ding{55}     & \cellcolor{yellow!25}18.12 & \cellcolor{yellow!25}0.512 & \cellcolor{yellow!25}0.417 & \cellcolor{yellow!25}0.164 \\
          & \ding{51}      & \ding{51}      & \ding{55}     & \ding{55}     & 16.87 & 0.463 &0.441 & 0.187 \\
          & \ding{51}      & \ding{55}     & \ding{55}     & \ding{51}      & \cellcolor{orange!25}18.88 & \cellcolor{orange!25}0.543 & \cellcolor{orange!25}0.391 & \cellcolor{orange!25}0.150 \\
          & \ding{51}      & \ding{51}      & \ding{55}     & \ding{51}      & \cellcolor{red!25}19.75 & \cellcolor{red!25}0.614 & \cellcolor{red!25}0.300   & \cellcolor{red!25}0.125 \\
    \bottomrule
    \end{tabular}%
}
    \vspace{-2mm}
    \caption{
    \textbf{Ablation Studies.} We perform ablation studies on Blender with 8 input views and LLFF with 3 input views, where \textbf{Pli} means per-layer inputs incorporation, \textbf{Sep} means separate modeling of colors and volume density, \textbf{Bkr} means background regularization, and \textbf{Sa} means sampling annealing.
    }
    \label{tab:ablation}
\end{table}

\begin{figure}
    \centering
    \begin{subfigure}[b]{1.0\linewidth}
         \centering 
    \includegraphics[width=\linewidth]
    {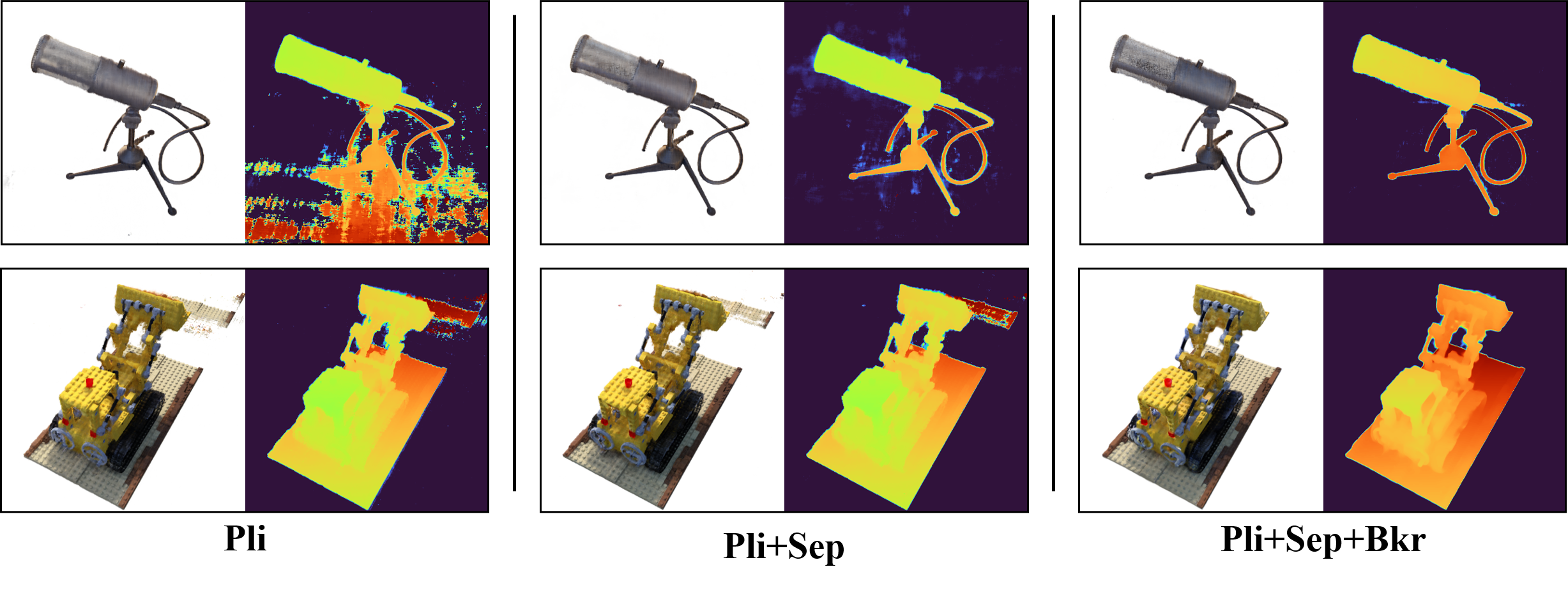}
    \vspace{-4mm}
    \caption{Qualitative results of ablation studies on Blender.}
    \end{subfigure}

    \begin{subfigure}[b]{1.0\linewidth}
         \centering 
    \includegraphics[width=\linewidth]
    {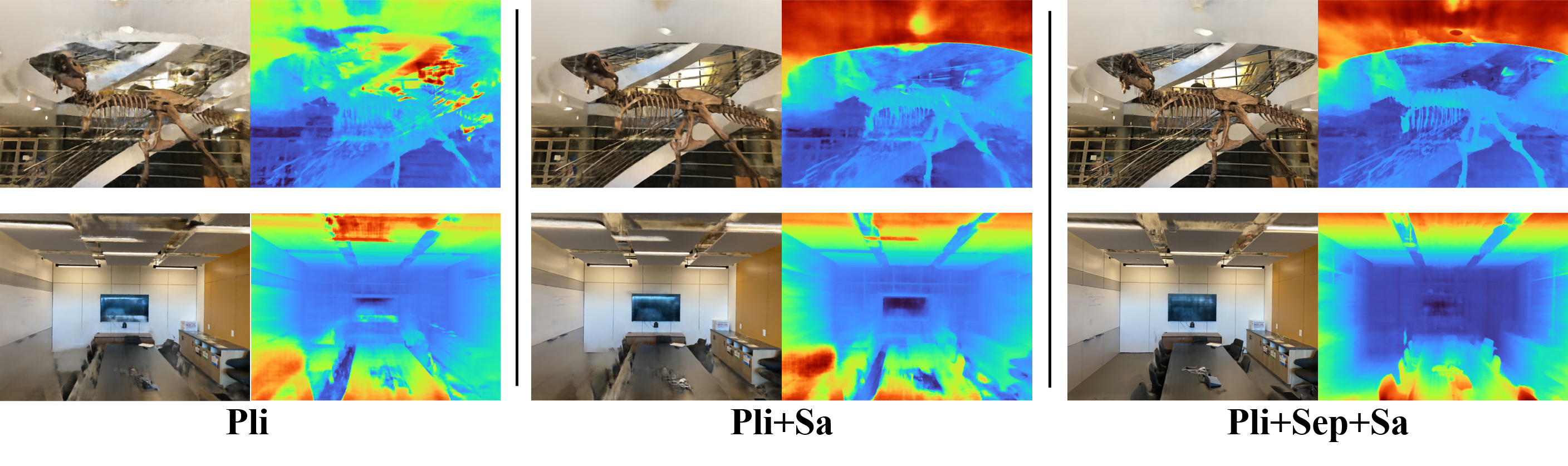}
    \vspace{-4mm}
    \caption{Qualitative results of ablation studies on LLFF.}
    \end{subfigure}
    \vspace{-6mm}
    \caption{Qualitative results of ablation studies on Blender and LLFF, where \textbf{Pli} means per-layer inputs incorporation, \textbf{Sep} means separate modeling of colors and volume density, \textbf{Bkr} means background regularization, and \textbf{Sa} means sampling annealing.}
    \vspace{-2mm}
    \label{fig:ablation_blender_8}
\end{figure}

\subsection{Ablation Studies}\label{sec:ablation}
To showcase the effectiveness of our design choices, we conduct both quantitative and qualitative ablation studies, as shown in Tab.~\ref{tab:ablation} and Fig.~\ref{fig:ablation_blender_8}. With only per-layer inputs incorporation, a dramatic performance gain against our baseline (\ie, vanilla NeRF) can be achieved, where we observe a $9.5$dB PSNR improvement for the Blender dataset and a $4.8$dB PSNR improvement for the LLFF dataset. For object-centric scenes like Blender, the separate modeling of volume density and colors is beneficial to clear geometry recovery, and the background regularization is able to further improve the performance by removing background artifacts. For forward-facing scenes like LLFF, we find that the sampling annealing strategy is crucial for accurate geometry estimation. By combining both the sampling annealing strategy and the separate modeling of volume density and colors, we are able to achieve the best performance. Moreover, we also try a classical approach to avoid overfitting, \ie, Dropout~\cite{srivastava2014dropout}, which we find a comparable performance with DietNeRF~\cite{jain2021putting} can be achieved. Kindly refer to the supplementary materials for more results.

\begin{table}
    \resizebox{\linewidth}{!}{\begin{tabular}{c|c|cccc}
    \toprule
    Method & Known Views & PSNR$\uparrow$  & SSIM$\uparrow$  & LPIPS$\downarrow$ & Average$\downarrow$ \\
    \midrule
    FreeNeRF~\cite{yang2023freenerf} & \multirow{2}{*}{4} & 18.88 & 0.777 & 0.179 & 0.102 \\
    FreeNeRF+Ours &       & \cellcolor{red!25}\textbf{19.36} & \cellcolor{red!25}\textbf{0.787} & \cellcolor{red!25}\textbf{0.173} & \cellcolor{red!25}\textbf{0.097} \\
    \midrule
    InfoNeRF~\cite{kim2022infonerf}& \multirow{2}{*}{8} & 24.27 & 0.868 & 0.112 & 0.053 \\
    InfoNeRF+Ours &       & \cellcolor{red!25}\textbf{24.77} & \cellcolor{red!25}\textbf{0.877} & \cellcolor{red!25}\textbf{0.104} & \cellcolor{red!25}\textbf{0.049} \\
    \midrule
    DietNeRF~\cite{jain2021putting} & \multirow{2}{*}{8} &   23.70    &  0.850     &   0.130    & 0.060 \\
    DietNeRF+Ours &       &   \cellcolor{red!25}\textbf{23.90}    &   \cellcolor{red!25}\textbf{0.857}    &   \cellcolor{red!25}\textbf{0.125 }  &  \cellcolor{red!25}\textbf{0.057}\\
    \bottomrule
    \end{tabular}%
}
    \vspace{-2mm}
    \caption{
    \textbf{Orthogonality of mi-MLP.} We choose 3 baselines, and replace their network structure with ours to demonstrate the proposed mi-MLP is orthogonal to current works.
    }
    \label{tab:compatibility}
\end{table}

\begin{figure}
    \centering
    \includegraphics[width=1\linewidth]{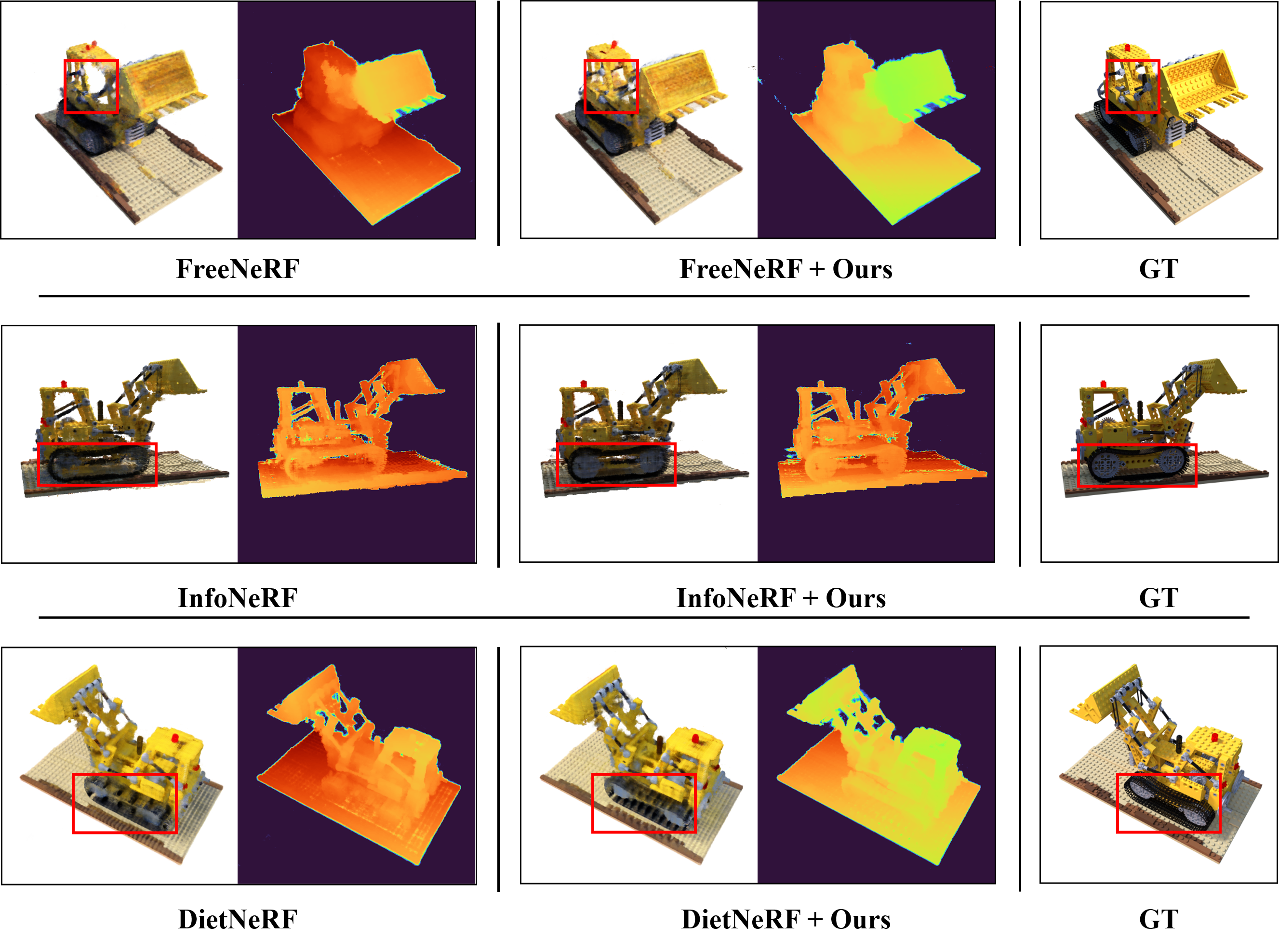}
    \vspace{-6mm}
    \caption{The proposed mi-MLP is orthogonal to current works since an improved performance can always be achieved for different methods when combined with our proposed mi-MLP.}
    \vspace{-2mm}
    \label{fig:Compatibility}
\end{figure}

\subsection{Orthogonality of mi-MLP}
We also perform experiments to demonstrate the proposed mi-MLP is orthogonal to current works. For this purpose, we select three representative methods: FreeNeRF~\cite{yang2023freenerf}, InfoNeRF~\cite{kim2022infonerf}, and DietNeRF~\cite{jain2021putting}, and replace their network structure with our proposed method.  As shown in Tab.~\ref{tab:compatibility} and Fig.~\ref{fig:Compatibility}, for a scene randomly chosen from the Blender dataset, better performance can always be achieved when combined with mi-MLP. Such a result reflects the potential of our proposed method to serve as a backbone for NeRF. Additionally, we extend mi-MLP to 3D generation, and the results are presented in the supplementary materials.

\section{Conclusion}
\vspace{-0.1cm}
In this paper, we have presented a novel method for few-shot view synthesis from the perspective of network structure for the first time. Specifically, to address the overfitting problem, motivated by the observation that a reduced model capacity is beneficial to alleviating overfitting while at the cost of missing details, we propose the mi-MLP that incorporates inputs into each layer of the MLP. Subsequently, based on the assumption that geometry is smoother than appearance, we propose to model colors and volume density separately for better geometry recovery. Additionally, we also provide two regularization terms to improve the quality of rendered novel views. Experiments have demonstrated that our proposed method can achieve state-of-the-art performance on multiple datasets. Considering the orthogonality of our proposed method, mi-MLP also opens up a new direction to other fields such as 3D generation. 
{
    \small
    \bibliographystyle{ieeenat_fullname}
    \bibliography{main}
}


\end{document}